# Some Complexity Considerations in the Combination of Belief Networks *

Izhar Matzkevich        Bruce Abramson
Computer Science Department and Social Science Research Institute
University of Southern California
Los Angeles, CA, 90089-0781
izhar@pollux.usc.edu
bda@pollux.usc.edu

## Abstract

One topic that is likely to attract an increasing amount of attention within the Knowledge-base systems resesearch community is the coordination of information provided by multiple experts. We envision a situation in which several experts independently encode information as belief networks. A potential user must then coordinate the conclusions and recommendations of these networks to derive some sort of consensus. One approach to such a consensus is the fusion of the contributed networks into a single, consensus model *prior* to the consideration of any case-specific data (specific observations, test results). This approach requires two types of combination procedures, one for probabilities, and one for graphs. Since the combination of probabilities is relatively well understood, the key barriers to this approach lie in the realm of graph theory. This paper provides formal definitions of some of the operations necessary to effect the necessary graphical combinations, and provides complexity analyses of these procedures. The paper's key result is that most of these operations are *NP*-hard, and its primary message is that the derivation of "good" consensus networks must be done heuristically.

## 1  INTRODUCTION

Thus far, the overwhelming majority of research on Knowledge-Base systems has been directed towards techniques for modeling domain information provided by a human expert, and for manipulating that information to yield insights into a specific problem instance within the domain. This research has led to several general frameworks for knowledge bases, including production rules, frames, formal logic, and belief networks (BN's). It has also helped raise several topics that promise to become increasingly important in the next wave of research. One such topic is the combination of multiple sources of expertise into a system that provides coherent recommendations based on a consensus of the contributing experts.

Our research focuses on the design of BN-based systems that combine several independently-designed BN's into a single system capable of providing consensus opinions and advice. In this paper, we consider some of the underlying theory necessary to design *prior compromise* networks, in which the combination of BN's occurs prior to the consideration of any case-specific data. Since BN's encode expertise through a combination of probability theory and graph theory, both numbers (probabilities) and structures (graphs) must be combined to yield a consensus BN. The combination of probabilities in the derivation of prior compromise is relatively well understood; Raiffa discussed the procedure's mechanics, its potential uses, and its merits relative to other methods for combining probabilities in 1968[8]. Our work concentrates on the fusion of graphical structures (which we consider as more fundamental) that is necessary to house these combined numbers. This paper provides a complexity-theoretic analysis of some of the tasks necessary to effect this structural combination.

## 2  THE GENERAL APPROACH

Most people (including experts) don't really see the world as a collection of formal models. They do, however, recognize that even within a specific domain, some items (or variables) are more closely interrelated than others. Observations of this sort lead to descriptions of dependence, indirect dependence, independence, and conditional (or partial) independence. As a result, it is often useful to think of the information provided by an expert as an abstract *independence model* [6]. Although many mathematical formalisms provide mechanisms for capturing independence, most of them only approximate these abstract independence models. Probability theory and graph theory—the two components of BN's—are two such formalisms. Pearl's development of the theory of BN's included proof that some forms of probabilistic independence can not be represented in a graph [6]. As a result, anything that is said about independencies captured in (and propagated through) a BN is only an approximation to the sorts of independencies that could be captured by (and manipulated in) an abstract independence model. In this paper, we demonstrate that some of the tasks necessary to combine two (or more generally, $k$) models into a single consensus model are intractable, whether they are performed within an abstract independence model or within a graphical representation (e.g., a BN).

---

*Supported in part by the National Science Foundation under grant SES-9106440.

The fundamental problem that we face arises because different contributors are likely to have different views of their domain of expertise. In particular, we expect to encounter disagreements about the interrelationships among variables, and specifically, about the representation of conditional independence among sets of variables[1]. As a result, some sort of polling mechanism must be used to provide a degree of confidence in each possible independence; confidence in an independence relation should be proportional to the number of contributors who claim it, and the consensus model should capture only independence relations in which we are confident. Therefore, given a set of $m \geq 2$ contributed models, we must determine some threshold $k$, $(1 \leq k \leq m)$, such that all independencies (and only those independencies) agreed upon by at least $k$ of the contributors are represented in the consensus model.

This paper's theoretical results indicate that modeling *all* independencies agreed upon by (all) subsets of contributors is impractical, even if there are only a few contributors, because the only way to do so would be to consider all possible (total) orderings of the domain' underlying variables. Furthermore, even if consideration is restricted to subsets of possible orderings (thereby sacrificing completeness), the problem of obtaining orderings that *maximize* the number of independencies preserved remains hard because the number of potential independencies on a given set of variables is exponential in the size of the set. This perceived intractability of even *optimizing* a consensus structure in an abstract independence model suggests that the problem might be easier in a more restrictive model, such as a graph. Since more arcs in a graph generally reduce the number of independencies captured (and thus increase the complexity of eliciting and manipulating information), a usable consensus model should *minimize* the number of arcs generated as a result of reordering a BN's underlying variables. This paper's main result is that even when the problem of maximizing the number of independencies that are captured in a consensus model is reduced to an optimization problem on DAGs (directed acyclic graphs, the structures used by BN's), the related optimization problems remain $NP$-hard.

This result has some significant implications. First, since an optimal efficient solution cannot be found in general, heuristic graphical methods are needed to solve the relevant DAG optimization problems. We have already presented one such algorithm for a related problem [4]; Shachter presented another [9]. The results presented in this paper provide additional justification for this type of approach.

## 3 PREVIOUS WORK

The basic groundwork upon which our results are based was laid by Pearl and his students in their development of the theory of BN's. The definitions and results presented in this section are taken (albeit with some minor modifications) from their work [5, 6, 10, 1].

A *dependency model* $M$ may be defined over a finite set of objects $U$ as any subset of triplets $(X, Y, Z)$ where $X, Y$ and $Z$ are three disjoint subsets of $U$. $M$ may be thought of as a truth assignment rule for the independence predicate, $I(X, Z, Y)$, read "$X$ is independent of $Y$, given $Z$" (an $I$-statement of this kind is called an *independency*, and its negation a *dependency*). An $I$-map of a dependency model $M$ is any dependency model $M'$ such that $M' \subseteq M$. A perfect map of a dependency model $M$ is any dependency model $M'$ such that $M' \subseteq M$ and $M \subseteq M'$.

**Definition 1** *A graphoid is any dependency model closed under the following axioms:*
*(i)* **Symmetry** $I(X, Z, Y) \Leftrightarrow I(Y, Z, X)$.
*(ii)* **Decomposition** $I(X, Z, Y \cup W) \Rightarrow I(X, Z, Y)$.
*(iii)* **Weak union** $I(X, Z, Y \cup W) \Rightarrow I(X, Z \cup W, Y)$.
*(iv)* **Contraction** $I(X, Z, Y)$ & $I(X, Z \cup Y, W) \Rightarrow I(X, Z, Y \cup W)$.
*A graphoid is intersectional if it also obeys the following axiom:*
*(v)* **Intersection** $I(X, Z \cup Y, W)$ & $I(X, Z \cup W, Y) \Rightarrow I(X, Z, Y \cup W)$.

Examples of graphoids include the *probabilistic dependency models* and the *acyclic digraph (DAG) models*. The criterion necessary for a DAG to capture an independence model is known as *d-separation*.

---
[1] A set of variables, A, is said to be conditionally independent of a second set, B, given a third set, C, if when C is unknown, information about B provides information about A, but when C is known, information about B provides no information about A. This definition is one of the most important concepts in the theory of BN's [6].

For any set $L$ of independencies, let $CL(L)$ denote $L$'s closure under the graphoid axioms.

In analyzing potential consensus structures, our aim was therefore to define graphical structures that capture (at least some of) the independencies represented in the input DAGs (assume without loss of generality that all are given over the same set of variables). Given $m \geq 2$ input BN's $B_i = \{V, \vec{E}_i, CP_i\}$, $1 \leq i \leq m$, let $D_i = (V, \vec{E}_i)$ be the DAG underlying $BN_i$, and $\alpha_i$ be a complete ordering on $V$ which is consistent with the partial ordering induced by $\vec{E}_i$. For each such $D_i$ then, define the set $L_{\alpha_i} = \{I(v, B_i(v), R_i(v)) | v \in V\}$, where for each $v \in V$, $B_i(v)$ is the set of immediate predecessors of $v$ in $D_i$, and $R_i(v)$ is the rest of the variables which precede $v$ in the ordering $\alpha_i$. $L_{\alpha_i}$ is termed the *recursive basis* drawn from $D_i$ relative to $\alpha_i$ [1]. These definitions led to the following two theorems [10]:

**Theorem 1** *For each $1 \leq i \leq m$, $CL(L_{\alpha_i})$ is a perfect map of $D_i$.*

In other words, $CL(L_{\alpha_i})$ captures every independency (and every dependency) that is graphically verified in $D_i$. $CL(L_{\alpha_i})$ is an intersectional graphoid. $CL(L_{\alpha_i})$ will therefore be used to denote the independencies captured by $D_i$ relative to the d-separation criterion.

**Theorem 2** *If a dependency model $M$ is a graphoid, then the set of DAGs generated from all recursive bases of $M$ is a perfect map of $M$ if the criterion for separation is that d-separation must exist in one of the DAGs.*

Theorems 1 and 2 therefore imply that *if $M$ is an intersectional graphoid, $A$ is the set of all total (complete) orderings on $M$'s variables, and for each $\alpha \in A$, $L_\alpha$ is the unique recursive basis drawn from $M$ relative to $\alpha$, then $\bigcup_{\alpha \in A} CL(L_\alpha) = M$*. Our analysis extends this basic result, and shows how it can be used to prove the difficulty of tasks related to the combination of BN's into a single consensus structure.

## 4  FURTHER THEORETICAL DEVELOPMENT

Recall that our analysis emerged from our desire to model the sets of independencies agreed upon by at least $k$ contributors. The results reviewed in the previous section deal only with the representational capabilities of a single model. We must therefore extend them to the point where they allow us to discuss both multiple models and the single model that emerges from their combination.

It is important to begin by noting that there are $\binom{m}{k}$ potential subsets of agreement among $k$ (out of $m$) input sources. Let $S_j$, $(1 \leq j \leq \binom{m}{k})$, denote each such subset of agreement, then the the requested set of independencies is: $\bigcup_{j=1}^{\binom{m}{k}} \bigcap_{i=1}^{k} CL(L_{\alpha_i}^{S_j})$, where for each $i,j$, $L_{\alpha_i}^{S_j} \in \{L_{\alpha_1}, \ldots, L_{\alpha_m}\}$. For example, given two input sources (using the above notation) the set of independencies agreed upon by at least *one* of the input sources (i.e., $m = 2, k = 1$) is $\bigcup_{i=1}^{2} CL(L_{\alpha_i})$, and the set of independencies that both of them agree upon is $\bigcap_{i=1}^{2} CL(L_{\alpha_i})$. Next, consider each such subset of $k \subseteq m$ input sources. Then:

**Lemma 1** *Given $1 \leq k \leq m$, then for any subset $S_j$, $1 \leq j \leq \binom{m}{k}$, of $k \subseteq m$ input sources, $\bigcap_{i=1}^{k} CL(L_{\alpha_i}^{S_j})$ is an (intersectional) graphoid (for each $i, j$ $L_{\alpha_i}^{S_j} \in \{L_{\alpha_1}, \ldots, L_{\alpha_m}\}$).*

Lemma 1 follows immediately because for each $1 \leq i \leq k$, $CL(L_{\alpha_i}^{S_j})$ is an intersectional graphoid. Now let $A$ be the set of all total (i.e., complete) orderings on $V$ (note that $|A| = |V|!$). For each $\alpha \in A$, $(1 \leq i \leq m)$, let $L_\alpha^i$ be the (unique) recursive basis drawn from $CL(L_{\alpha_i})$ (i.e., $D_i$), relative to $\alpha$. For each subset $S_j$, $1 \leq j \leq \binom{m}{k}$, of $k \subseteq m$ input sources, and $\alpha \in A$, define a *k-unified-recursive-basis*, $L_\alpha^{S_j}$, as $I(v, \bigcup_{i=1}^{k} B_i(v), \bigcap_{i=1}^{k} R_i(v)) \in L_\alpha^{S_j} \Leftrightarrow \bigwedge_{i=1}^{k}[I(v, B_i(v), R_i(v)) \in L_\alpha^{S_j,i}]$, where $L_\alpha^{S_j,i}$ is the recursive basis drawn from $CL(L_{\alpha_i}^{S_j})$ relative to the total ordering $\alpha$. This definition leads to the following lemma:

**Lemma 2** *Given $1 \leq k \leq m$, $\alpha \in A$, and a subset $S_j$, $1 \leq j \leq \binom{m}{k}$, of $k \subseteq m$ input sources, then, (i) $L_\alpha^{S_j}$ is exactly the recursive basis drawn from $\bigcap_{i=1}^{k} CL(L_{\alpha_i}^{S_j})$ relative to the total ordering $\alpha$, and (ii) $CL(L_\alpha^{S_j})$ is a minimal I-map of $\bigcap_{i=1}^{k} CL(L_\alpha^{S_j,i})$.*

To further clarify lemmas 1 and 2 (as well as their implications), consider once again the example of two input models $D_1 = (V, \vec{E}_1), D_2 = (V, \vec{E}_2)$ (i.e., $CL(L_{\alpha_1})$, $CL(L_{\alpha_2})$). Given $k = 2$ and some $\alpha \in A$, let $L_\alpha^i$, $i = 1, 2$ be the recursive bases drawn from $CL(L_{\alpha_i})$ relative to $\alpha$ (note that there is only one such a subset when $k = m$, and hence the superscript $S_1$ is omitted). Lemma 1 implies that $\bigcap_{i=1}^{2} CL(L_{\alpha_i})$ (the set of independencies agreed upon by both input sources) is an intersectional graphoid. Lemma 2 implies that $L_\alpha$, the 2-unified-recursive-basis drawn relative to $\alpha$, is a one such that $CL(L_\alpha)$ is a minimal I-map of $\bigcap_{i=1}^{2} CL(L_\alpha^i)$. Now, let $D_\alpha^i$, $i = 1, 2$ be the DAG generated by $L_\alpha^i$ (for each $v \in V$, point an arc from each $u \in B_i(v)$ to $v$), then $D_\alpha$, the DAG generated by $L_\alpha$, is not only the *union-DAG* of $D_\alpha^1$ and $D_\alpha^2$ (i.e., if $D_\alpha^i = (V, \vec{E}_i), i = 1, 2$ then $D_\alpha = (V, \bigcup_{i=1}^2 \vec{E}_i)$), but also a minimal $I$-map of $\bigcap_{i=1}^2 CL(L_\alpha^i)$ relative to the $d$-separation criterion (in general though, it is not a perfect map of it).

Lemmas 1 and 2 combine with the following theorem to establish a formal justification for using union-DAGs to represent a consensus by integrating sets of independencies agreed upon by any subset of $k$ of the $m$ input sources into a single structure [4].

**Theorem 3** *For any $k$, $1 \le k \le m$,*

$$\bigcup_{\alpha \in A} \bigcup_{j=1}^{\binom{m}{k}} CL(L_\alpha^{S_j}) = \bigcup_{\alpha \in A} \bigcup_{j=1}^{\binom{m}{k}} \bigcap_{i=1}^k CL(L_\alpha^{S_j,i})$$
$$= \bigcup_{j=1}^{\binom{m}{k}} \bigcap_{i=1}^k CL(L_{\alpha_i}^{S_j}).$$

A proof of theorem 3 follows (inductively) from theorem 2 (which corresponds the special case of $k = m = 1$), and lemmas 1 and 2. Theorem 3 thus implies that the collection of all $k$-union-DAGs, for all $\alpha \in A$, and $S_j$, $1 \le j \le \binom{m}{k}$, forms a perfect map (relative to $d$-separation) for the set of independencies agreed upon by at least $k$ of the $m$ input sources. The problem, of course, is that although this result may be meaningful from a theoretical standpoint, it is of no practical value when $V$ is sufficiently large (even for $k = m = 2$).

Theorem 3 holds when the input models are intersectional graphoids. It is fairly simple to show that in general, a perfect coverage can not be derived when union is taken over only a *polynomial* (in $|V|$) number of closures of recursive bases, even when the input models are DAGs (for which closure properties other than the intersectional graphoid axioms hold [6]). One reasonable fall-back then, might be to derive an ordering $\alpha$ for which the number of (non-trivial) independencies (all are assumed to be of an equal 'importance' at this point) 'captured' by some $\bigcap_{i=1}^k CL(L_\alpha^{S_j,i})$ (i.e., $CL(L_\alpha^{S_j})$) is *maximized* (i.e., for all $\alpha' \in A$, $|CL(L_{\alpha'}^{S_j})| \le |CL(L_\alpha^{S_j})|$). This type of a compromise is reasonable since for each $\alpha \in A$, the DAG induced by $L_\alpha^{S_j}$ is a *minimal $I$-map* (relative to $d$-separation) of $\bigcap_{i=1}^k CL(L_{\alpha_i}^{S_j})$ (i.e., no arc can be removed without destroying the $I$-mapness property).

Now, given any $\alpha \in A$, and a set $S_j$, deriving $L_\alpha^{S_j,i}$, $1 \le i \le k$ (and hence $L_\alpha^{S_j}$) is rather straightforward. Our attempt to focus on orderings that *maximize* $|CL(L_\alpha^{S_j})|$, however, is considerably harder for two reasons. First, all possible orderings over the underlying set of variables should somehow be considered. Second, there may be $O(4^n)$ potential nontrivial independencies over $n$ variables (a result easily obtained using the multinomial theorem). Readers familiar with the problem should probably notice by now that the notion of *entailment* among belief networks (for which graphical criteria were presented in [7]) is closely related; it is yet unclear how can it be applied in our case.

The accurate and efficient identification of an ordering $\alpha \in A$ which induces a recursive basis (or equivalently, deriving the recursive basis itself), whose closure under the graphoid axioms is of maximal cardinality, is left as an open problem. Instead, we resort to a *heuristic graphical approach* in order to maximize the number of independencies captured by a consensus model. An *arc-reversal* operation applied over a DAG $D = (V, \vec{E})$ [9], may generate new arcs; when an arc $(u, v) \in \vec{E}$ is reversed, new arcs may be generated from each vertex in $P_D(u) \setminus P_D(v)$ to $v$, and from each vertex in $P_D(v) \setminus P_D(u)$ to $u$ (for each $v' \in V$, $P_D(v')$ is the set of immediate predecessors of $v'$ in $D$). Each such new arc induces new dependencies, and thereby eliminates some independencies captured by $D$ (relative to $d$-separation). Assume now a subset $S_j$ of size $k$, $1 \le k \le m$, of the $m$ input models is given. If each of $S_j$'s members is to be "rearranged" when the relevant recursive bases are derived relative to some total ordering $\alpha$, sequences of arc-reversals might be required on (some of) the them. This is a process by which independencies may be eliminated. Therefore, in order to maximize the number of independencies captured by the recursive bases' intersection relative to any ordering $\alpha$ (i.e., maximizing $|CL(L_\alpha^{S_j})|$ over all $\alpha \in A$), one would wish to identify a total ordering for which, for example, the sequences of arc-reversals required are of minimal lengths. Alternatively, one would wish to identify a total ordering that minimizes the number of arcs generated as a result of applying arc-reversals on the relevant DAGs.

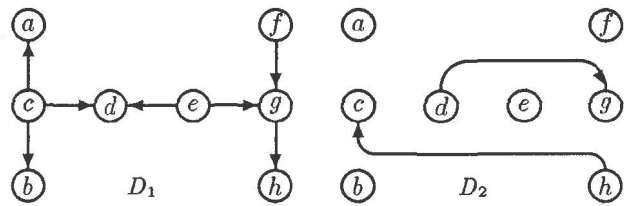

Figure 1: Minimizing the number of newly generated arcs does not guarantee maximal number of independencies.

Figure 1 demonstrates that topological optimizations do not guarantee optimality in capturing indepen-dencies, and are thus only heuristics. In this example, $D_1$ is to be rearranged, so that the relevant union-DAG is acyclic. Assume that arc $(g, h)$ is reversed. This causes the creation of two new arcs $(e, h)$ and $(f, h)$. Reversing the arc $(c, d)$, on the other hand, creates only arc $(e, c)$. Although two new arcs are generated when reversing $(h, g)$, compared to only one when reversing $(c, d)$, it appears that the number of independencies lost in the latter reversal is *larger*. The following section formally assesses the complexity of some of these optimization problems.

## 5 COMPLEXITY ANALYSIS

The following discussion is limited to the case of two input sources. The crux of the analysis is a demonstration that several of the operations necessary to combine independence models, even those represented as DAGs, are *NP*-hard.

Our analysis begins with a graph theoretic problem known to be *NP*-complete, the *minimum feedback arc set problem* (**FAS**) [3, 2]: *Given a digraph* $D = (V, \vec{E})$ *and a positive integer* $k$, *is there a subset of arcs* $\vec{E}'$, *such that* $|\vec{E}'| \leq k$ *and* $D' = (V, \vec{E} \setminus \vec{E}')$ *is acyclic?* We consider an optimization variant of **FAS**, called **MFAS**: *Given such a digraph* $D = (V, \vec{E})$, *find a minimal such a set* $\vec{E}'$ *(which need not be unique) such that* $(V, \vec{E} \setminus \vec{E}')$ *is acyclic.* In this context, "minimal" means: such that for any other set $\vec{E}''$ for which $(V, \vec{E} \setminus \vec{E}'')$ is acyclic, $|\vec{E}'| \leq |\vec{E}''|$. It is clear that **FAS** $\propto_p$ **MFAS**, (where the operation $\propto_p$ denotes a polynomial time reduction from one problem to another). Thus, **MFAS**, like many other optimization variants of *NP*-complete problems [2], is *NP*-hard.

Next, we define a problem called **MRS**: *Given a digraph* $D = (V, \vec{E})$, *find a minimal such a set* $\vec{E}'$ *such that if* $\vec{E}'_R = \{(u, v) | (v, u) \in \vec{E}'\}$, *then* $(V, (\vec{E} \setminus \vec{E}') \cup \vec{E}'_R)$ *is acyclic.* **MRS**, like **MFAS**, looks for a minimal set $\vec{E}' \subseteq \vec{E}$ such that reversing $\vec{E}'$-s arcs renders the resulted digraph acyclic.

**Theorem 4** : *MRS is* NP-*hard*.

**Sketch of proof.** We show that **MFAS** $\propto_p$ **MRS**. This by showing that for any *minimal* set $\vec{E}' \subseteq \vec{E}$, $D'' = (V, (\vec{E} \setminus \vec{E}') \cup \vec{E}'_R)$ is acyclic iff $D' = (V, \vec{E} \setminus \vec{E}')$ is. The *if* part is obvious. The *only-if* part, however, requires the following claim:

**Claim 1** *If* $\vec{E}'$ *is a minimal set such that* $D' = (V, \vec{E} \setminus \vec{E}')$ *is acyclic, then for each* $(u, v) \in \vec{E}'$, *there exists at least one such directed cycle (denote its set of arcs* $C_{(u,v)}$) *in* $D$, *for which* $\vec{E}' \cap C_{(u,v)} = \{(u,v)\}$ *(i.e.,* $C_{(u,v)}$ *is 'exclusive' for* $(u, v)$ *in that sense).*

Claim 1 implies that if $D'$ is acyclic, but $D''$ is not, then examining each such directed cycle $C$ formed in $D''$ as a result of reversing $\vec{E}'$, let $\vec{E}'_C = \vec{E}' \cap C$, then $(C \setminus \vec{E}'_C) \cup \bigcup_{(u,v) \in \vec{E}'_C} C_{(u,v)}$ is a directed cycle which must also exist in $D'$ (thereby rendering it *cyclic*, a contradiction), or else we must violate the *minimality* of $\vec{E}'$. (Each such $C_{(u,v)}$, $(u,v) \in \vec{E}'_C$ is one such an 'exclusive' $(u, v)$-directed-cycle guaranteed by claim 1). □

Polynomial reductions combine in a transitive form. Therefore, since **MRS** is *NP*-hard, it is now possible to show that the set of optimization problems that interest us are all *NP*-hard. Recall that our aim—given 2 input input BN's, with $D_1 = (V, \vec{E}_1)$, $D_2 = (V, \vec{E}_2)$ their underlying acyclic digraph—is to construct a *union* acyclic digraph $D$ (relative to some total ordering $\alpha$ on $V$) over $D_1, D_2$ such that the number of independencies captured by $D$ (relative to $d$-separation), is *maximized*. For this problem, we noted that applying a sequence of arc-reversal operations [9] on (each of) the digraphs, a sequence which *minimizes* the number of arcs generated in $D_1$ and $D_2$ as a result, is a reasonable heuristic. Such a sequence of arc reversal operations rearranges the input digraphs so that the partial ordering imposed by $\vec{E}_i$ on $V$ in $D_i$, $i = 1, 2$ is consistent with $\alpha$.

As we are about to show, however, this optimization procedure (problem) is *NP*-hard. Moreover, even in its simplified form, when the target total ordering $\alpha$ is a one which is consistent with one of the input acyclic digraphs, say $D_2$ (whereby only $D_1$ is "rearranged" if necessary)—it still is *NP*-hard. This simplified version of the problem is where we start the complexity analysis.

Now, let $D_1, D_2$ be defined as above, and examine the following problem, **DMRS**: *Find a* minimal *set* $\vec{E}' \subseteq \vec{E}$, *such that the digraph* $D = (V, (\vec{E}_1 \setminus \vec{E}') \cup \vec{E}'_R \cup \vec{E}_2)$ *is acyclic.*

**Theorem 5** : *DMRS is* NP-*hard*.

**Sketch of proof.** We show that **MRS** $\propto_p$ **DMRS**. Given $D = (V, \vec{E})$ an instance of **MRS**, for each $(u, v) \in \vec{E}$ we define a set of vertices $V_{(u,v)}$ such that $|V_{(u,v)}| = 2$, $V \cap V_{(u,v)} = \emptyset$, and for each $(u, v), (u', v') \in \vec{E}, (u, v) \neq (u', v') \Rightarrow V_{(u,v)} \cap V_{(u',v')} = \emptyset$ (i.e., each such $V_{(u,v)}$ contains two *unique* symbols). Next, for each $(u, v) \in \vec{E}$, let $V_{(u,v)} = \{u_{(u,v)}, v_{(u,v)}\}$, we define the following sets of arcs $\vec{E}^1_{(u,v)} = \{(u_{(u,v)}, v_{(u,v)})\}$, $\vec{E}^2_{(u,v)} = \{(u, u_{(u,v)}), (v_{(u,v)}, v)\}$. Then, by taking $\vec{E}_1 = \bigcup_{(u,v) \in \vec{E}} \vec{E}^1_{(u,v)}$, $\vec{E}_2 = \bigcup_{(u,v) \in \vec{E}} \vec{E}^2_{(u,v)}$ (intuitively, each arc in $\vec{E}$ is 'broken' into 3 parts, of which the center one is in $\vec{E}_1$, and the other two in $\vec{E}_2$). Then, we define the following two acyclic digraphs $D_1 = (V \cup \bigcup_{(u,v) \in \vec{E}} V_{(u,v)}, \vec{E}_1)$, $D_2 = (V \cup \bigcup_{(u,v) \in \vec{E}} V_{(u,v)}, \vec{E}_2)$ as an instance of **DMRS**. Thus, $D' = (V, (\vec{E} \setminus \vec{E}') \cup \vec{E}'_R)$ is acyclic iff $D'' = (V \cup \bigcup_{(u,v) \in \vec{E}} V_{(u,v)}, (\vec{E}_1 \setminus \vec{E}'') \cup \vec{E}''_R \cup \vec{E}_2)$ is, where $(u,v) \in \vec{E}' \Leftrightarrow (u_{(u,v)}, v_{(u,v)}) \in \vec{E}''$.    □

This construction further indicates that the problem of minimizing the number of arc-reversal operations performed on $D_1$ is at least as complex. (In fact, this problem could be rephrased: *find a minimal sequence of arc-reversal operations on $D_1$ such that the union-digraph $D'_1 \cup D_2$ is acyclic, where $D'_1$ is the digraph obtained from $D_1$ by applying this sequence of arc-reversals*).

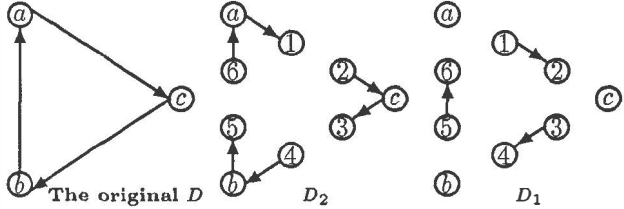

Figure 2: The reduction of **MRS** into **DMRS**.

Next, we define **2DMRS**: *Let $D_1 = (V, \vec{E}_1)$, $D_2 = (V, \vec{E}_2)$ be two acyclic digraphs, then find a minimal set $\vec{E}' \subseteq \vec{E}_1 \cup \vec{E}_2$, such that the digraph $D = (V, (\vec{E}_1 \cup \vec{E}_2) \setminus \vec{E}' \cup \vec{E}'_R)$ is acyclic*. In this case, therefore, reversals are allowed in both $D_1, D_2$.

**Theorem 6** : **2DMRS** *is* NP-*hard*.

**Sketch of a proof.** We show that **DMRS** $\propto_p$ **2DMRS**. Given two acyclic digraphs $D_1 = (V, \vec{E}_1)$, $D_2 = (V, \vec{E}_2)$ as an instance of **DMRS**, then for each $(u,v) \in \vec{E}_2$, define a set of vertices $V_{(u,v)}$ such that $|V_{(u,v)}| = |V|^2$, $V \cap V_{(u,v)} = \emptyset$, and moreover $V_{(u,v)} \cap V_{(u',v')} = \emptyset \Leftrightarrow (u,v) \neq (u',v')$. Next, for each such $(u,v) \in \vec{E}_2$, define the following set of arcs $\vec{E}_{(u,v)} = \{(u,v'),(v',v) | v' \in V_{(u,v)}\}$ (intuitively, we replace each arc $(u,v) \in \vec{E}_2$ by a set of $|V|^2$ pairs of arcs $(u,v'),(v',v)$). Finally, define the acyclic digraph $D'_2 = (V \cup \bigcup_{(u,v) \in \vec{E}_2} V_{(u,v)}, \bigcup_{(u,v) \in \vec{E}_2} \vec{E}_{(u,v)})$. Taking $D_1$, $D'_2$ as an instance of **2DMRS**, it is readily seen that any minimal set $\vec{E}' \subseteq \vec{E}_1 \cup \bigcup_{(u,v) \in \vec{E}_2} \vec{E}_{(u,v)}$ reversed is a one such that $\vec{E}' \subseteq \vec{E}_1$, and furthermore, is exactly the minimal set $\vec{E}'$ required for the **DMRS** instance under hand.    □

This construction, combined with the one given for **MRS** $\propto_p$ **DMRS**, further implies that the more general problem of minimizing the number of arc-reversal operations, when such reversals are allowed on both input digraphs, is *NP*-hard as well.

Finally, consider **MNAS**: *Given two acyclic digraphs $D_1 = (V, \vec{E}_1)$, $D_2 = (V, \vec{E}_2)$, find a sequence of arc reversals on $D_1$, such that the union-digraph $D'_1 \cup D_2$ is acyclic, $D'_1 = (V, \vec{E}'_1)$ is the digraph obtained from $D_1$ by applying this sequence of arc-reversals, and furthermore, the set $\vec{E}'_1 \setminus (\vec{E}_1 \cup (\vec{E}_1)_R)$ is minimal (i.e., the set of new arcs generated as a result of 're-arranging' $D_1$, is minimal)*. Recall that minimizing this set of new arcs is a heuristics that we apply towards maximizing the number of independencies captured by $D'_1 \cup D_2$ relative to the *d*-separation criterion.

**Theorem 7** : **MNAS** *is* NP-*hard*.

**Sketch of a proof.** We show that **MRS** $\propto_p$ **MNAS**. In fact, the reduction mechanism is very similar to the one used in showing **MRS** $\propto_p$ **DMRS**. Given $D = (V, \vec{E})$ an instance of **MRS**, for each $(u,v) \in \vec{E}$ we define a set of vertices $V_{(u,v)}$ such that $|V_{(u,v)}| = 3$, $V \cap V_{(u,v)} = \emptyset$, and for each $(u,v),(u',v') \in \vec{E}, (u,v) \neq (u',v') \Rightarrow V_{(u,v)} \cap V_{(u',v')} = \emptyset$ (i.e., each such $V_{(u,v)}$ contains 3 *unique* symbols). Next, for each $(u,v) \in \vec{E}$, let $V_{(u,v)} = \{u_{(u,v)}, v_{(u,v)}, w_{(u,v)}\}$, we define the following sets of arcs
$\vec{E}^1_{(u,v)} = \{(u_{(u,v)}, v_{(u,v)}),(w_{(u,v)}, v_{(u,v)})\}$, $\vec{E}^2_{(u,v)} = \{(u, u_{(u,v)}),(v_{(u,v)}, v)\}$. Then, by taking $\vec{E}_1 = \bigcup_{(u,v) \in \vec{E}} \vec{E}^1_{(u,v)}$, $\vec{E}_2 = \bigcup_{(u,v) \in \vec{E}} \vec{E}^2_{(u,v)}$, (i.e., $|\vec{E}_2| = |\vec{E}_1| = 2|\vec{E}|$), we define the following two acyclic digraphs $D_1 = (V \cup \bigcup_{(u,v) \in \vec{E}} V_{(u,v)}, \vec{E}_1)$, $D_2 = (V \cup \bigcup_{(u,v) \in \vec{E}} V_{(u,v)}, \vec{E}_2)$ as an instance of **MNAS**.

Given such $D_1$, $D_2$, and a sequence $S$ of arc-reversals that minimizes the number of newly generated arcs, let $\vec{E}'$ be that minimal set of *new* arcs generated as a result of applying $S$ on $D_1$, then deriving the set $\vec{E}''$ requested by **MRS** on $D$, it is clearly seen that $(u,v) \in \vec{E}'' \Leftrightarrow (u_{(u,v)}, w_{(u,v)})$ or $(w_{(u,v)}, u_{(u,v)}) \in \vec{E}'$, where $u_{(u,v)}, w_{(u,v)} \in V_{(u,v)}$.    □

Now consider a related problem, **2MNAS**, of finding a sequence of arc-reversals which minimizes the number of newly generated arcs, this time allowing arc-reversal operations on *both* the input digraphs.

**Theorem 8** : **2MNAS** *is* NP-*hard*.

**Sketch of a proof.** We show that **MNAS** $\propto_p$ **2MNAS**. Given two input acyclic digraphs $D_1 = (V, \vec{E}_1)$, $D_2 = (V, \vec{E}_2)$ as an instance of **MNAS**, for each $u \in V$ define the following set of vertices $V_u$ such that $|V_u| = |V|^2$, $V \cap V_u = \emptyset$, and moreover $V_u \cap V_{u'} = \emptyset \Leftrightarrow u \neq u'$. Next, de-fine $\vec{E}'_2 = \vec{E}_2 \cup \bigcup_{u \in V}\{(u', u) | u' \in V_u\}$ (i.e., with each $u \in V$ we introduce a unique set of $|V|^2$ new arcs), and consider the two acyclic digraphs $D'_1 = (V \cup \bigcup_{u \in V} V_u, \vec{E}_1)$, $D'_2 = (V \cup \bigcup_{u \in V} V_u, \vec{E}'_2)$. Taking $D'_1, D'_2$ as an instance of **2MNAS**, it is readily seen that for minimizing the number of newly generated arcs, arc-reversals should only be performed in $D'_1$. Moreover, any minimizing sequence of arc-reversals on $D'_1$—along with the resulted set $\vec{E}'$ of newly generated arcs—is also a minimizing sequence of arc reversals on $D_1$ with the exact same $\vec{E}'$ as the requested minimal set. □

## 6   SUMMARY

Although this paper was highly theoretical, the fundamental issues that it addressed grew out of a practical concern: our desire to develop BN-based systems that incorporate the input of several contributing experts. That practical objective led us to identify two distinct subproblems, the combination of numbers (i.e., probabilities), and the combination of structures (i.e., graphs). Since the combination of probabilities is relatively well understood, we decided to focus on structural combination. This decision, in turn, led us to consider the rather theoretical problem of combining abstract independence models into a single consensus model. Since one important early step in algorithm design is an analysis of the underlying complexity of the tasks being tackled, we turned our attention to a complexity-theoretic analysis of some of the operations necessary to combine independence models. Although not analyzed in this paper, it appears that virtually all of these operations are *NP*-hard. We therefore turned our attention to graphical models, which are, of course, only an approximation of abstract independence models, and we showed that here too, most of the operations needed to generate "optimal" consensus structures are *NP*-hard. We have, however, already been able to demonstrate that the generation of consensus structures is both doable and tractable [4]. We thus end this paper with a simple conclusion: the generation of "good" consensus structures (of the type necessary to generate tractable consensus BN's) will require the use of heuristics. These heuristics should probably be based on a combination of the domain being modeled and the topology of the contributed models. Research on this topic is currently underway. We hope that it will lead to not only an elegant theory of consensus BN's, but also to a practical, applicable procedure that helps combine the contributions of multiple experts into a coherent consensus-based system.